\documentclass[letterpaper, 10 pt, conference]{ieeeconf}  % Comment this line out if you need a4paper

\IEEEoverridecommandlockouts                              % This command is only needed if 
\usepackage{amsmath} % assumes amsmath package installed
\usepackage{amssymb}  % assumes amsmath package installed
\usepackage{amsthm}
\usepackage{amsfonts}

\usepackage{graphicx}
\usepackage{array}
\usepackage{subcaption}
\usepackage{bm}
\usepackage{bbm}
\usepackage{algorithm2e}
\RestyleAlgo{ruled}
\SetAlCapNameFnt{\small}
\SetAlCapFnt{\small}
\usepackage{algpseudocode}
\usepackage{multirow}
\usepackage{float}
\usepackage{placeins}
\newcommand\norm[1]{\left\lVert#1\right\rVert}
\newcommand{\1}{\mbox{\fontencoding{U}\fontfamily{bbold}\selectfont1}}
\newcommand{\0}{\mbox{\fontencoding{U}\fontfamily{bbold}\selectfont0}}

\DeclareMathAlphabet{\mathcal}{OMS}{cmsy}{m}{n}
\newcommand{\Prob}{\mathbb{P}}
\newcommand{\E}{\mathbb{E}}

\usepackage{hyperref}%
\hypersetup{colorlinks=true, linkcolor=blue, breaklinks=true, urlcolor=blue}

\renewcommand{\natural}{{\mathbb{N}}}

\newcommand{\real}{\ensuremath{\mathbb{R}}}

\newcommand{\tauvec}{\pmb{\tau}}
\newcommand{\pivec}{\pmb{\pi}}

\newcommand{\e}{\bm{e}}
\newcommand{\M}{\mathcal{M}}

\newcommand{\diag}{\textup{diag}}
\newcommand{\vecvec}{\textup{vec}}

\title{\LARGE \bf RoSSO: A High-Performance Python Package for \underline{Ro}botic \underline{S}urveillance \underline{S}trategy \underline{O}ptimization Using JAX}
\author{Yohan John, Connor Hughes, Gilberto D\'iaz-Garc\'ia, Jason R. Marden, and Francesco Bullo% <-this % stops a space
\thanks{This work was supported by Air Force Office of Scientific Research under Grant FA9550-15-1-0138.}% <-this % stops a space
\thanks{The authors thank Sean Jaffe and Max Emerick for valuable discussions in the early stages of this work.}%
\thanks{The authors are with the Center for Control, Dynamical Systems, and Computation, UC Santa Barbara, Santa Barbara, CA 93106-5070, USA. Email: \{{\tt\small yohanjohn, connorhughes, gdiaz-garcia, jrmarden, bullo}\}{\tt\small@ucsb.edu}}
}

\begin{document}

\maketitle
\thispagestyle{empty}
\pagestyle{empty}

%%%%%%%%%%%%%%%%%%%%%%%%%%%%%%%%%%%%%%%%%%%%%%%%%%%%%%%%%%%%%%%%%%%%%%%%%%%%%%%%
\begin{abstract}
To enable the computation of effective randomized patrol routes for single- or multi-robot teams, we present RoSSO, a Python package designed for solving Markov chain optimization problems. We exploit machine-learning techniques such as reverse-mode automatic differentiation and constraint parametrization to achieve superior efficiency compared to general-purpose nonlinear programming solvers. Additionally, we supplement a game-theoretic stochastic surveillance formulation in the literature with a novel greedy algorithm and multi-robot extension. We close with numerical results for a police district in downtown San Francisco that demonstrate RoSSO's capabilities on our new formulations and the prior work. 
\end{abstract}
%%%%%%%%%%%%%%%%%%%%%%%%%%%%%%%%%%%%%%%%%%%%%%%%%%%%%%%%%%%%%%%%%%%%%%%%%%%%%%%%

\section{Introduction}

% \subsection{Motivation \& Problem Description}
Conducting surveillance with autonomous mobile robots offers the potential for reduced risk in dangerous environments. However, deterministic patrol routes can be exploited by intelligent adversaries. As a result, randomized patrol routing has gained popularity. Unfortunately, identifying optimal stochastic patrol strategies for realistic environments is often intractable. In this article, we present RoSSO~\cite{CH-YJ:23}, a Python package for efficiently handling these problems via first-order methods.  

\subsection{Related Work}
Robotic surveillance has been discussed in the literature for over two decades~\cite{PER-al:00}. A common approach is to discretize the environment into a graph and then design patrol routes on the graph to deter or capture intruders~\cite{SMLV-DL-LJG-JCL-RM:97}. Early on, it was recognized that stochastic patrols can be more effective than deterministic patrols~\cite{JPH-HJK-SS:99}. Markov chains (MCs) are a natural choice for representing a stochastic patrol strategy~\cite{JG-JB:05}, and priority among the nodes can be encoded in the stationary distribution~\cite{RP-PA-FB:14b}. 
% We refer the reader to~\cite{XD-FB:20a} for a review of MC-based robotic surveillance.

Several metrics have been proposed for identifying the ``best" MC. An early work~\cite{SB-PD-LX:04} proposed the fastest mixing Markov chain (FMMC), but was confined to a special class of MCs for convexity. More recently, minimizing the mean first hitting time of an MC was explored and showed improved performance over FMMC~\cite{RP-PA-FB:14b}. Once again, a constraint was added for convexity. These approaches focus on the speed of the robotic patroller. In~\cite{MG-SJ-FB:17b}, the authors proposed maximizing the entropy rate, a classical measure of unpredictability, of an MC. This formulation is convex, but does not account for travel times on the edges of the graph. Alternatively, maximizing the return-time entropy leads to unpredictable return intervals to the nodes in the graph~\cite{XD-MG-FB:17o}. Unfortunately, this formulation does not admit a closed-form and must be approximated. 
% A projected gradient ascent algorithm is presented for optimizing the approximation numerically. 
Most recently, a Stackelberg game formulation was proposed for maximizing the probability of detecting an omniscient adversary~\cite{XD-DP-FB:19b}. Finding the optimal patrol strategy in this formulation is a nonconvex optimization problem.
% \jason{this paragraph is very bullo-centric.  would be nice to diversify..} 

In summary, globally optimizing a patrol strategy MC is intractable, regardless of the choice of metric. A practical approach then is local optimization via RoSSO. 
% The metrics discussed previously are built-in and implementing new metrics is straightforward. 
The most closely related open-source software is implemented in MATLAB and Julia~\cite{HW:08-github}. 
% However, RoSSO has more capabilities built-in and is more efficient due to its implementation in Python with JAX. 
However, RoSSO provides more capabilities, includes new problem formulations, and is more efficient due to its implementation in Python with JAX. 

\subsection{Contributions}
RoSSO leverages tools from the machine-learning community to provide a means of computing effective MC patrol strategies. Specifically, RoSSO can accommodate arbitrary graph topology, heterogeneous edge travel times, and varying node priority along with the choice of metric. These capabilities enable RoSSO to be applied for real-world patrol design. We highlight the following contributions:
% \jason{would be nice if there was a more complete statement of the problem up to this point.}
\begin{enumerate}
    \item an efficient Python package containing a JAX-based gradient optimizer for several MC metrics,
    \item a novel greedy algorithm for co-optimizing defense placement and patrol strategy in the Stackelberg game formulation,
    \item a novel multi-robot Stackelberg game formulation with an efficient method for handling a stationary distribution constraint, and
    \item numerical results demonstrating these contributions with RoSSO on a police district in San Francisco.
\end{enumerate}
Additionally, RoSSO can serve as a useful tool for education and research. It provides a modular framework that can be easily extended for a variety of real-world considerations including a large number of robot patrollers~\cite{GDG-FB-JRM:22v}, vision limitations~\cite{TA-MMR-PC-LB-BP:20}, battery life constraints~\cite{SH-AR-HD:20}, etc. 

\section{Problem Formulation} \label{sect:problem_form}

\subsection{Notation}
Let $\real, \natural$ be the sets of real and natural numbers, respectively. Let $\1_n \in \real^n$ be the vector of all ones, and $\0_n \in \real^n$ be the vector of all zeros. The $i$-th basis vector is denoted $\e_i$. 
% $\diag(A)$ denotes a diagonal matrix with the same entries on the diagonal as the matrix $A$. 
$\diag(\bm{a})$ denotes a diagonal matrix whose diagonal entries are taken from the vector $\bm{a}$. Let $\vecvec(\cdot)$ be the column-wise vectorization operator. $\circ, \otimes$ are the Hadamard and Kronecker products, respectively.
% \jason{this section all starts off with preliminaries and less about model.  Do you want to have a preliminaries section?  Seems awkward. } 

\subsection{Graphs \& Markov Chains}
Consider a patrol environment graph $\mathcal{G} = (\mathcal{V},\mathcal{E}, W)$ where $\mathcal{V}$ is the node set, $\mathcal{E}$ is the edge set, and $W$ is a weight matrix. Nodes represent patrol waypoints, edges represent patrol paths, and the $(i,j)$-th entry of $W$ is the integer travel time between waypoint $i$ and waypoint $j$. 
% If $(i,j) \notin \mathcal{E}$, then we define $W(i, j) = 0$.
Note that the unit of travel time is arbitrary so we can rescale to achieve any level of precision. 
% However, the computational expense increases with temporal resolution. 
We use a discrete-time MC to represent the stochastic patrol strategy for this graph. An $n$-state MC, for an $n$-node graph, can be encapsulated in a row-stochastic transition matrix $P\in \real^{n\times n}$. 
The $(i,j)$-th entry of $P$, $P(i,j)$, is the probability of the robot patroller taking the $(i,j)$ path when departing from node $i$. Additional background on the stationary distribution and first hitting times of MCs is presented in the Appendix.
% Should we mention that irreducibility is a necessary condition for optimality for SG?
The following three sections will review the MC metrics that we will evaluate with RoSSO.

\subsection{Mean Hitting Time (MHT) Formulation} \label{sect:mht}
Here we briefly review the MHT formulation proposed in~\cite{RP-PA-FB:14b}. This formulation aims to identify the fastest-moving randomized patrol strategy for a given graph. The approach is to minimize the expectation of the first hitting time $T_{ij}$ for all pairs of nodes $(i,j)$. The MHT $\M \in \real$ is defined as $\M = \pivec^\top M \pivec$ where $\pivec$ is the stationary distribution and $M \in \real^{n \times n}$ is the matrix whose $(i,j)$-th entry is $\E[T_{ij}]$. 
% $\M$ can also be calculated as $\M = 1 + \sum_{j=2}^n 1 / (1 - \lambda_j)$ where the $\lambda_j$'s are the eigenvalues of $P$ and $\lambda_1 = 1$. 
% If there are heterogeneous travel times on the edges of the graph, the weighted MHT $\M^w \in \real$ can be computed as:
We focus on the weighted MHT objective function $J_{\textup{MHT}}$ that accommodates heterogeneous travel times on the edges of the graph
\begin{equation}
    J_{\textup{MHT}}(P) = (\pivec^\top(P \circ W)\1_n)\M(P).
\end{equation}

\subsection{Return-Time Entropy (RTE) Formulation} \label{sect:rte}
In this section, we recap the RTE formulation proposed in~\cite{XD-MG-FB:17o}. Consider an attacker who discreetly observes his target location for a period of time prior to attacking. If the patroller arrives at relatively predictable intervals, then the attacker can act accordingly. To prevent this, the RTE formulation identifies the patrol strategy with the most unpredictable return time to each node in the graph. Unfortunately, the RTE does not admit a closed-form; therefore, we maximize the truncated RTE defined as
\begin{equation} \label{eq:ret_ent}
    J_{\textup{RTE}}(P) = -\sum\nolimits_{i=1}^n \pi_i \sum\nolimits_{k=1}^{K_\eta} F_k(i,i) \log F_k(i,i),
\end{equation}
where $F_k$ is the $k$-th first hitting time probability matrix and $K_{\eta} = \lceil w_{\textup{max}} / (\eta \pi_{\textup{min}}) \rceil - 1$. $\eta \in (0,1)$ is a truncation accuracy parameter that upper bounds the discarded probability.

\subsection{Stackelberg Game (SG) Formulation} \label{sect:sg}
Here we summarize the SG formulation proposed in~\cite{XD-DP-FB:19b} and extended in~\cite{YJ-GDG-XD-JRM-FB:23}. This game-theoretic formulation identifies the optimal patrol strategy for a patroller facing an omniscient attacker. 
% The omniscient attacker is assumed to know the location of the patroller and the patrol strategy. 
The attacker has to remain stationary at a node $i$ for a given duration $\tau_i \in \natural$ in order to complete an attack and win the game. If the patroller visits that node within $\tau_i$ time periods, then the patroller wins. This can be written as a max-min problem to identify the optimal $P^*$:
\begin{equation} \label{eq:Stack}
    J_{\textup{SG}}(P^*) = \max_{P} \min_{i,j} \bigl\{ \Prob[T_{ij}(P) \leq \tau_j] \bigr\}.
\end{equation}
The inner minimization reflects the attacker's choice of a node $j$ to attack while the patroller is at node $i$. The outer maximization reflects the patroller's desire to maximize the probability of visiting the attacker's node via the choice of $P$. The probabilities of the patroller visiting node $j$ from node $i$ within $\tau_j$ time periods can be collected into a capture probability matrix $\Lambda = \sum_{j=1}^n \sum_{k=1}^{\tau_j} F_k \e_j \e_j^\top \in \real^{n\times n}$. The patroller's goal is to maximize the worst-case capture probability, i.e., the minimum entry of $\Lambda$. 

\section{RoSSO}
In this section, we present the key features of the RoSSO package. Additional details can be found in the documentation provided with the codebase on GitHub~\cite{CH-YJ:23}.
% \subsection{Repository Structure}
% RoSSO is structured as recommended by Kenneth Reitz~\cite{KR:13}. The subdirectory `RoSSO' contains the code which is split among six Python files. \texttt{graph\_gen.py} and \texttt{graph\_comp.py} handle creation and manipulation of graphs. \texttt{test\_spec.py} contains class code for \texttt{TestSpec} objects that specify the desired inputs and parameters for a particular suite of optimization runs. \texttt{strat\_comp.py} and \texttt{strat\_opt.py} evaluate and optimize a given patrol strategy, respectively. Finally, \texttt{strat\_viz.py} provides methods for visualizing the results.
RoSSO contains six Python modules. The modules \texttt{graph\_gen.py} and \texttt{graph\_comp.py} handle creation and manipulation of the environment graphs. The \texttt{test\_spec.py} module defines the \texttt{TestSpec} class, which provides useful infrastructure for defining a suite of optimization runs. \texttt{strat\_comp.py} contains objective function definitions, gradient computations, and constraint parametrization. The strategy optimization algorithm and related performance tracking features are implemented in \texttt{strat\_opt.py}. Finally, \texttt{strat\_viz.py} provides a variety of methods for visualizing surveillance strategies and optimization metrics.
% \texttt{TestSpec} objects are used to load, save, modify, and validate the environment graphs, attack durations, and optimizer parameters that define a suite of optimization runs. 

\subsection{JAX} 
RoSSO utilizes JAX and Optax for gradient-based optimization of surveillance strategies. JAX is a library for machine-learning research which enables automatic differentiation and just-in-time compilation via TensorFlow's accelerated linear algebra compiler~\cite{JB-RF-PH-MJJ-CL-DM-GN-AP-JVP-SWM-QZ:18}. JAX seamlessly facilitates acceleration on GPU/TPU hardware while providing a familiar NumPy-like interface. Optax builds upon JAX and offers additional tools including a host of optimization algorithms from the machine-learning literature~\cite{deepmind2020jax}. RoSSO leverages these features to provide a simple yet powerful optimization framework for the robotic surveillance community. Because of its modular architecture, RoSSO is readily extended to handle new problem formulations simply by defining new objective functions or constraints. 

% \subsection{Optimization Constraints}
% \subsection{Constraints and Objective Functions}
\subsection{Handling Constraints}
There are four constraints on the transition matrix $P$: graph constraints, nonnegativity, row-stochasticity, and a given stationary distribution. The first three constraints are mandatory for a valid MC subordinate to a graph. The stationary distribution constraint is used to encode varying priority among the nodes, and its inclusion is optional. 
We handle the mandatory constraints via a parametrization function that takes an arbitrary matrix $Q \in \real^{n \times n}$ and returns a valid transition matrix $P$:
\begin{enumerate}
    \item Graph constraint: $P \gets Q \circ A$
    \item Nonnegativity: $P \gets |P|$
    \item Row-stochasticity: $P(i,:) \gets P(i,:) / \sum_{j=1}^n P(i,j), \forall i$
\end{enumerate}
where the absolute value $|\cdot|$ is applied element-wise and $A$ is the binary adjacency matrix of the graph. In RoSSO, the constraint parametrization is composed with each objective function defined in Section~\ref{sect:problem_form} prior to automatic differentiation, so gradient-based updates are applied in $Q$-space. Valid optimized strategies are then obtained by evaluating the parametrization function with the final $Q$. 
% In this way, RoSSO leverages JAX's flexibility to handle new problem formulations with minimal modification to the codebase.
% The gradients are propagated through the parametrization steps to make updates in $Q$-space. 

The stationary distribution constraint is handled by augmenting the objective function $J$, which could be any of the metrics discussed in Section~\ref{sect:problem_form}, with a penalty term:
% If the stationary distribution constraint is included, we handle it by augmenting the objective function $J$ with a least-squares penalty term:
\begin{equation}\label{eq:penalty}
    \min_P J(P) + \alpha \norm{\pivec^\top P - \pivec^\top}_2^2,
\end{equation}
where $\alpha > 0$ is a tunable hyperparameter. 
% Choosing a large value of $\alpha$ enforces the stationary distribution constraint exactly while a smaller value allows some deviation in order to decrease $J$. 
For maximization problems we subtract the same penalty term. 

% QUESTION: should we state that including the constraint parametrization in the objective function we autodiff makes this a policy-gradient method? Need to verify this, heard it from Sean Jaffe. 

\subsection{Stackelberg Game Co-Optimization Formulation}\label{sect:coopt}
In this section, we present a novel algorithm for a variant of the SG formulation described in Section~\ref{sect:sg}. As proposed in~\cite{YJ-GDG-XD-JRM-FB:23}, the vector of attack durations $\tauvec = \begin{bmatrix}    \tau_1 & \cdots & \tau_n
\end{bmatrix}$ in Eq.~\eqref{eq:Stack} can be viewed as a measure of the strength of the defenses at each node in the graph. We consider the simultaneous optimization of patrol strategy $P$ and defense placement $\tauvec$, given a defense budget $B$. This can be written:
\begin{subequations} \label{eq:co_opt}
\begin{align}
    J_{\textup{SGC}}(P^*, \tauvec^*) & = \max_{P,\tauvec} \min_{i,j} \bigl\{ \Prob[T_{ij}(P) \leq \tau_j] \bigr\} \\
    \text{s.t.} \quad & P \ \textup{row-stochastic}, \\
     \quad & \sum\nolimits_{i=1}^n \tau_i = B, \\
    \quad & \tau_i \in \natural, \quad \forall i.
\end{align}
\end{subequations}
In~\cite{YJ-GDG-XD-JRM-FB:23}, approximation algorithms for~\eqref{eq:co_opt} are presented for certain types of graphs with homogeneous travel times. We want RoSSO to be applicable to arbitrary graph topologies and heterogeneous travel times. However,~\eqref{eq:co_opt} is ill-suited for a gradient-based optimization approach because the $\tau_i$ decision variables are integer-valued. Therefore, we propose the greedy defense placement algorithm, Alg.~\ref{alg:greedy_def_alloc}, for choosing $\tauvec$ which can then be composed  with RoSSO's efficient gradient-based optimization of the patrol strategy $P$. The algorithm reflects the intuitive approach of iteratively increasing the level of defense at the node with the lowest capture probability until the budget is exhausted.
\begin{algorithm}
\caption{Greedy defense placement}\label{alg:greedy_def_alloc}
Given $P, B$\;
Use $P$ to compute $F_k$ matrices via Eq.~\eqref{eq:F_k_homo} or Eq.~\eqref{eq:F_k_het} for $k \in \{1, \ldots, B-n+1\}$\;
$M \gets \0_{n \times n}, \quad \tauvec \gets \0_n$\;
\While{$B > 0$}{
    Identify column $j$ containing min entry of $M$\;
    Increment corresponding entry of $\tauvec$: $\tau_j \gets \tau_j + 1$\;
    Update the $j$-th column of $M$: $M(:,j) \gets M(:,j) + F_{\tau_j}(:,j)$\;
    $B \gets B - 1$\;
}
Return $\tauvec, M$
\end{algorithm}

\subsection{Multi-Robot Stackelberg Game Formulation}\label{sect:multi}
Here we present a novel multi-robot formulation in the SG framework. The major obstacle in the multi-robot patrolling problem is the curse of dimensionality. Prior work has focused on either partitioning the graph and assigning one robot to each subgraph~\cite{GDG-FB-JRM:22v} or identifying a cyclic patrol strategy and spacing robots evenly along it~\cite{YC:04}. 
The cyclic strategy is known to be effective for some graphs and can be randomized~\cite{NA-SK-GAK:08}, but there are difficulties in maintaining even spacing in hardware~\cite{NA-CLF-YE-PS-CJ-SV:12}.
% The cyclic strategy is known to be effective and can be randomized~\cite{NA-SK-GAK:08}, however, maintaining even spacing with real hardware presents challenging communication requirements~\cite{NA-CLF-YE-PS-CJ-SV:12}.
We aim to design and evaluate effective non-cyclic stochastic patrol strategies for multiple patrolling robots. 

% With the appropriate computing hardware, RoSSO is efficient enough to handle a few robots on a small graph, and we present those results here. 
We consider a team of $R$ patrolling robots where each robot moves around a graph $\mathcal{G}^r = (\mathcal{V},\mathcal{E}^r,W^r)$ following an MC patrol strategy $P^r$. Note that the node set $\mathcal{V}$ is common to all the robots, but the edge set $\mathcal{E}^r$ and travel times $W^r$ can be unique to each robot $r$. This formulation accommodates heterogeneous robot teams such as a combination of multirotors, legged robots, wheeled robots, etc. The robot team's first hitting time $T_{i_1 \dots i_R,j}$ is defined as the number of time periods until at least one of the robots visits node $j$~\cite{RP-AC-FB:14k}:
\begin{equation}
\begin{aligned}
    T_{i_1 \dots i_R,j} = \min \{ k \geq 1 | & X_0^1 = i_1, \dots, X_0^R = i_R, \\
    & X_k^1 = j\ \textup{or} \ \cdots \ \textup{or} \ X_k^R = j \}.
\end{aligned}
\end{equation}

We propose a multi-robot extension to the SG formulation. 
% We will show in Section~\ref{sect:results} that the SG formulation provides a balance of speed and unpredictability preferable to the MHT and RTE formulations.
% The return-time entropy formulation in~\eqref{eq:ret_ent} is too computationally expensive to extend due to the need to calculate $F_k$ matrices for $k \in \{1,\dots, N_{\eta}\}$.
The first step is computing the first hitting time matrices $F_k^r$ for each robot independently for $k \in \{1,\ldots,\tau_\textup{max} \}$ using Eq.~\eqref{eq:F_k_homo} or Eq.~\eqref{eq:F_k_het}. Then the capture probability can be defined as the probability of at least one robot visiting the attacker's node within the attack duration (which is equal to one minus the probability of none of the robots capturing the attacker):
\begin{equation}
\begin{aligned}
    & \Prob[T^1_{i_1j} \leq \tau_j \ \textup{or} \ \cdots \ \textup{or} \ T^R_{i_Rj} \leq \tau_j] \\
    & = 1 - \prod_{r=1}^R (1 - \Prob[T^r_{i_rj} \leq \tau_j]) \\
    & = 1 - \prod_{r=1}^R \Bigl[ 1 - \sum\nolimits_{k=1}^{\tau_j} F_k^r(i_r, j) \Bigr].
\end{aligned}
\end{equation}
The capture probabilities can be collected into a tall matrix $\Lambda \in \real^{n^R \times n}$ where each row corresponds to an initial configuration for the robot team and each column corresponds to a potential node for the attacker. The objective function $J_{\textup{SGM}}$ becomes maximizing the minimum entry of $\Lambda$. Note that the $n^{R+1}$ possible configurations for the robot team and the attacker can be pre-computed and stored. However, this is still a computationally intensive formulation which is only feasible for a small number of robots. The efficiency of RoSSO allows us to present first-of-their-kind multi-robot results on a realistic graph in Section~\ref{sect:res_multi}.

Inspired by~\cite{RP-AC-FB:14k}, we enforce a stationary distribution constraint on a team of robots via an objective function penalty on the average stationary distribution $\pivec_{\textup{avg}} = (1/R) \sum_{i=1}^R \pivec^i$:
\begin{equation}\label{eq:multi_penalty}
    \max_{P^1,\dots,P^R} J_{\textup{SGM}}(P^1,\dots,P^R) - \alpha \norm{\pivec_{\textup{avg}} - \pivec}_2^2,
\end{equation}
where $\pivec$ is the desired stationary distribution and $\pivec^i$ are the individual robot's stationary distributions. To evaluate this objective function for a given set of strategies $P^1,\ldots,P^R$, we need to be able to efficiently compute the stationary distributions $\pivec^1,\ldots,\pivec^R$. Because the stationary distribution is the left eigenvector corresponding to the largest magnitude eigenvalue of the transition matrix, we propose Alg.~\ref{alg:power_iter}, a power iteration method based on~\cite{RVM-HPG:29} for efficiently computing $\pivec^i$. Note that $\pivec, P$ are nonnegative so the norm in the update step can simply be the sum.
\begin{algorithm}
\caption{Power iteration for stationary distribution computation}\label{alg:power_iter}
Given $P^i,\textup{max\_iter}$\;
Set $\pivec^i \gets (1/n)\1_n,\ \textup{iter} \gets 0$\;
\While{$\textup{iter} < \textup{max\_iter}$}{
    Update $\pivec^i \gets (\pivec^i)^\top P^i / \norm{(\pivec^i)^\top P^i}$\;
    $\textup{iter} \gets \textup{iter} + 1$\;
}
Return $\pivec^i$
\end{algorithm}

\section{Results} \label{sect:results}
\begin{figure}
    \centering
    \includegraphics[width=0.35\textwidth]{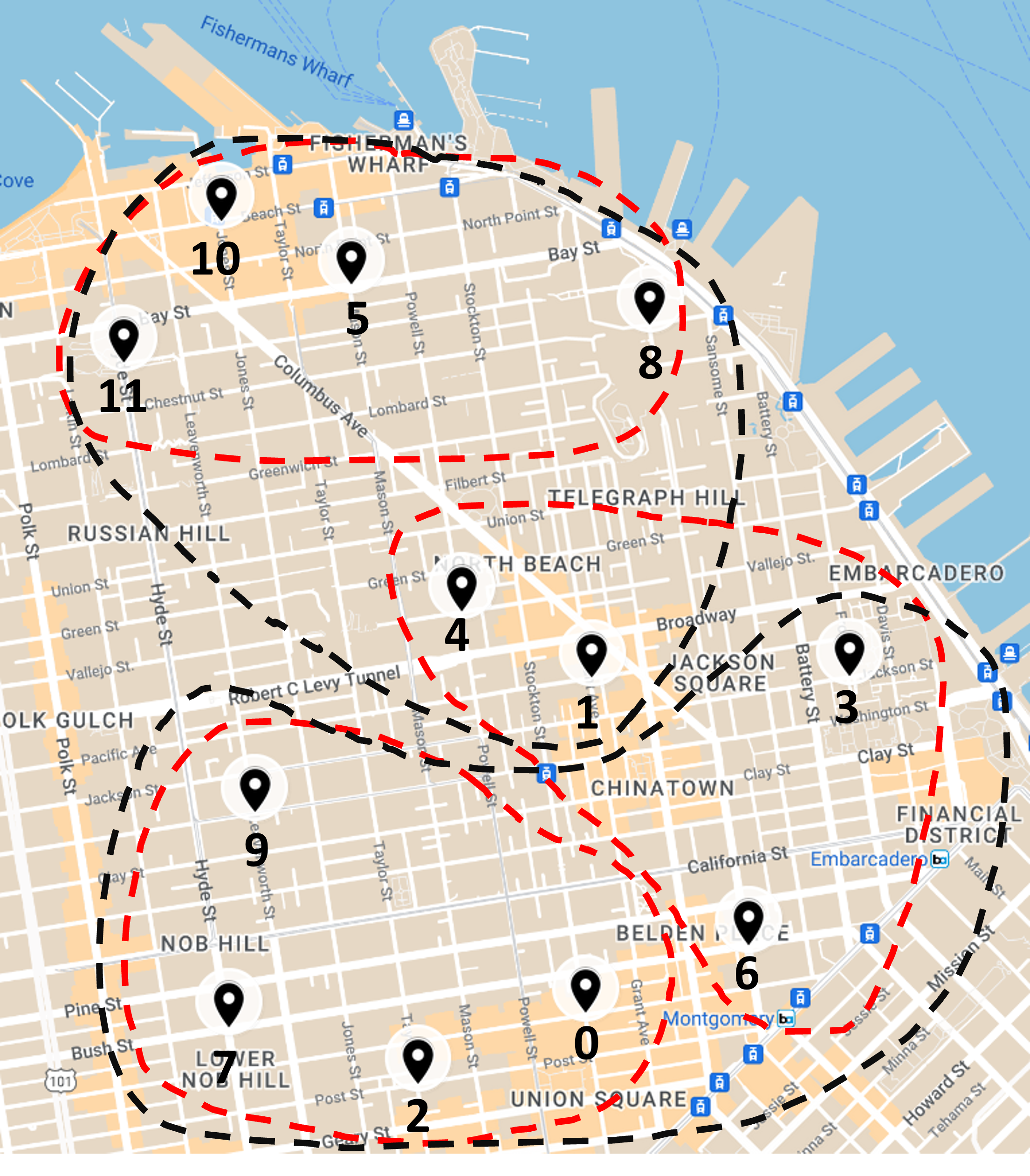}
    \caption{Twelve intersections in the central district of the San Francisco police department.}
    % The dashed black and red regions represent partitions for a team of $N = 2$ and $ N = 3$ robot patrollers, respectively.
    \label{fig:SF}
\end{figure}

In this section, we present the case study of a police district in downtown San Francisco. The environment is modeled as 12-node complete graph, see Fig.~\ref{fig:SF}~\cite{SA-EF-SLS:14}. The nodes represent important intersections in the district. The partitions shown by the closed dashed curves will be discussed in Section~\ref{sect:res_multi}. We include the stationary distribution constraint with $\pivec = \tfrac{1}{866}[\begin{smallmatrix} 133 & 90 & 89 & 87 & 83 & 83 & 74 & 64 & 48 & 43 & 38 & 34 \end{smallmatrix}]^\top$ chosen to be proportional to the monthly crime rates at each intersection~\cite{SA-EF-SLS:14}. The travel time matrix $W$ is given by the following where the values represent driving time in minutes~\cite{SA-EF-SLS:14}:
\begin{equation} \setcounter{MaxMatrixCols}{20}
\begin{aligned}
    W & = \tiny \begin{bmatrix}
        1 & 3 & 3 & 5 & 4 & 6 & 3 & 5 & 7 & 4 & 6 & 6 \\
        3 & 1 & 5 & 4 & 2 & 4 & 4 & 5 & 5 & 3 & 5 & 5 \\
        3 & 5 & 1 & 7 & 6 & 8 & 3 & 4 & 9 & 4 & 8 & 7 \\
        6 & 4 & 7 & 1 & 5 & 6 & 4 & 7 & 5 & 6 & 6 & 7 \\
        4 & 3 & 6 & 5 & 1 & 3 & 5 & 5 & 6 & 3 & 4 & 4 \\
        6 & 4 & 8 & 5 & 3 & 1 & 6 & 7 & 3 & 6 & 2 & 3 \\
        2 & 5 & 3 & 5 & 6 & 7 & 1 & 5 & 7 & 5 & 7 & 8 \\
        3 & 5 & 2 & 7 & 6 & 7 & 3 & 1 & 9 & 3 & 7 & 5 \\
        8 & 6 & 9 & 4 & 6 & 4 & 6 & 9 & 1 & 8 & 5 & 7 \\
        4 & 3 & 4 & 6 & 3 & 5 & 5 & 3 & 7 & 1 & 5 & 3 \\
        6 & 4 & 8 & 6 & 4 & 2 & 6 & 6 & 4 & 5 & 1 & 3 \\
        6 & 4 & 6 & 6 & 3 & 3 & 6 & 4 & 5 & 3 & 2 & 1
    \end{bmatrix}.
\end{aligned}
\end{equation}
Results were obtained using a laptop with an Intel Core i7 CPU (1.61 GHz) and 16 GB RAM.

\subsection{Single-Robot Patrolling}
First we study single-robot patrolling. We begin by substantiating the claim that RoSSO is more computationally efficient than the existing software~\cite{HW:08-github}. We compare with their MATLAB implementation that leverages the general-purpose constrained nonlinear programming solver \texttt{fmincon}. Table~\ref{table:HW} contains the comparison data averaged over 10 runs. The SG formulation is excluded from the comparison because~\cite{HW:08-github} does not appear to yield meaningful results for that formulation. The wall times are comparable for the MHT formulation, but the optimized objective function value for~\cite{HW:08-github} is noticeably worse due to their inclusion of the reversibility constraint on the MC. We see a dramatic speedup using RoSSO for the RTE formulation. We expect the performance improvements of RoSSO to increase on GPU/TPU hardware due to implementation in JAX.
\begin{table}
\centering
\caption{Computational efficiency comparison between RoSSO and~\cite{HW:08-github} averaged over 10 random initializations.}
\label{table:HW}
\begin{tabular}{| c | c | c c |} 
 \hline
  \multicolumn{2}{|c|}{} & MHT & RTE \\
 \hline
 \multirow{2}{8em}{RoSSO} & Avg. Wall Time [s] & 1.14 & 11.5 \\ 
  & Avg. Obj. Fun. Value & 23.7 & 5.00 \\ 
  \hline
 \multirow{2}{8em}{\cite{HW:08-github} (MATLAB)} & Avg. Wall Time [s] & 0.81 & 409 \\ 
  & Avg. Obj. Fun. Value & 44.8 & 5.00 \\ 
 \hline
\end{tabular}
\end{table}

\begin{figure*}
    \centering
     \begin{subfigure}[b]{0.24\textwidth}
         \centering
         \includegraphics[width=\textwidth]{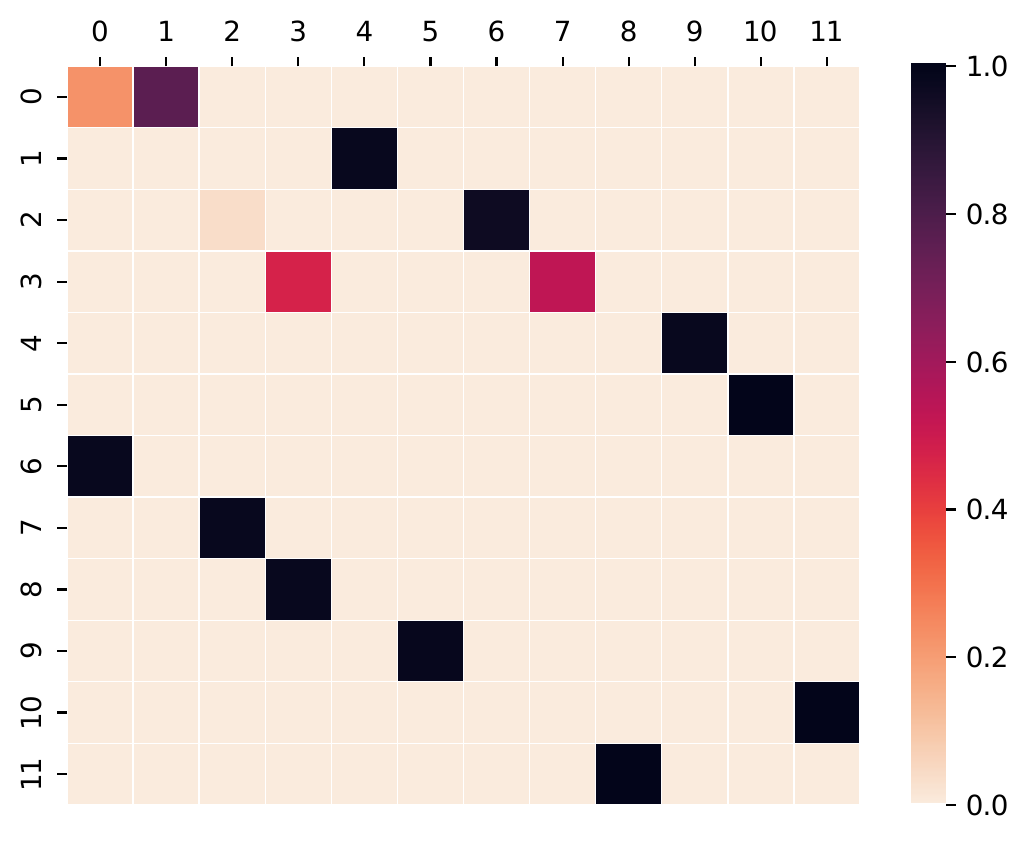}
         \caption{MHT, Section~\ref{sect:mht}}
         \label{fig:MHT}
     \end{subfigure}
     \hfill
     \begin{subfigure}[b]{0.24\textwidth}
         \centering
         \includegraphics[width=\textwidth]{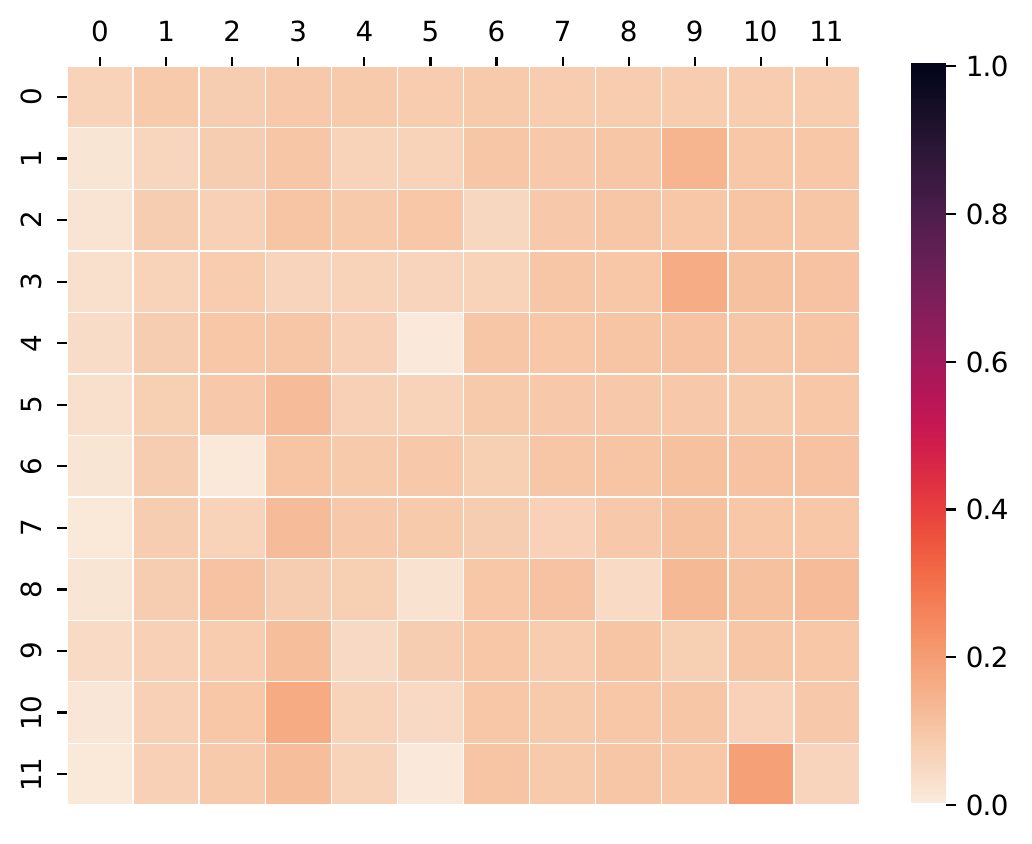}
         \caption{RTE, Section~\ref{sect:rte}}
         \label{fig:RTE}
     \end{subfigure}
     \hfill
     \begin{subfigure}[b]{0.24\textwidth}
         \centering
         \includegraphics[width=\textwidth]{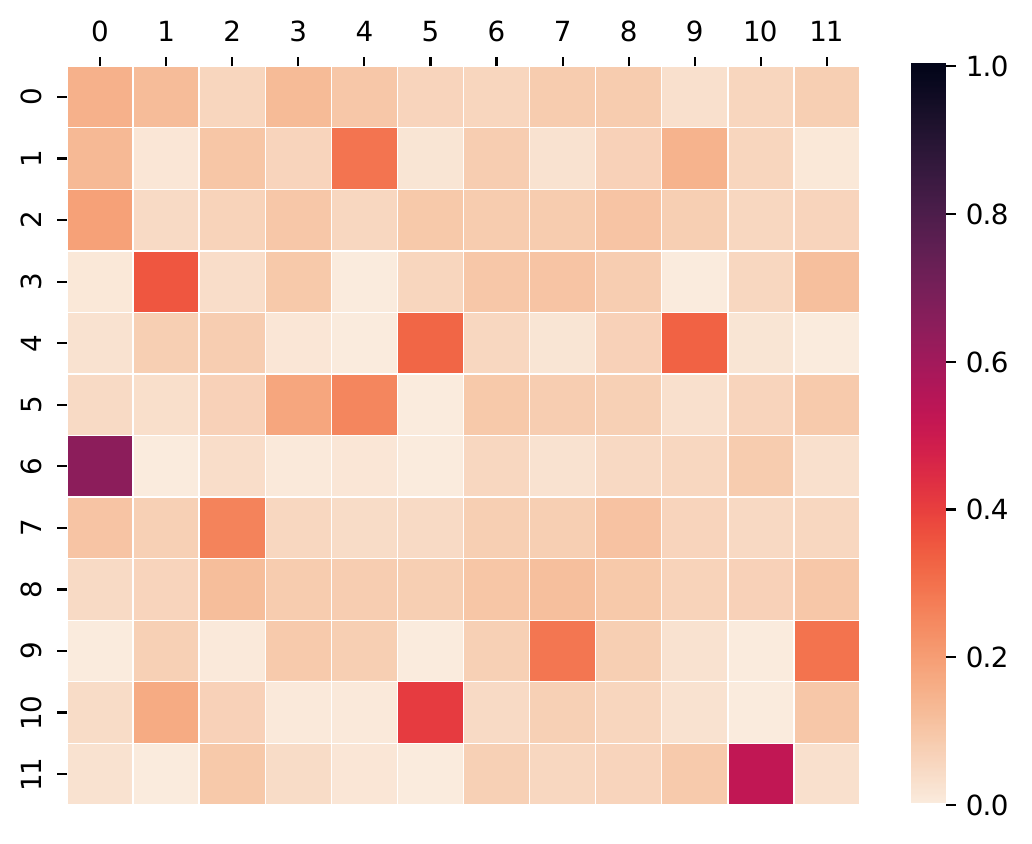}
         \caption{SG, Section~\ref{sect:sg}}
         \label{fig:SG}
     \end{subfigure}
     \hfill
     \begin{subfigure}[b]{0.24\textwidth}
         \centering
         \includegraphics[width=\textwidth]{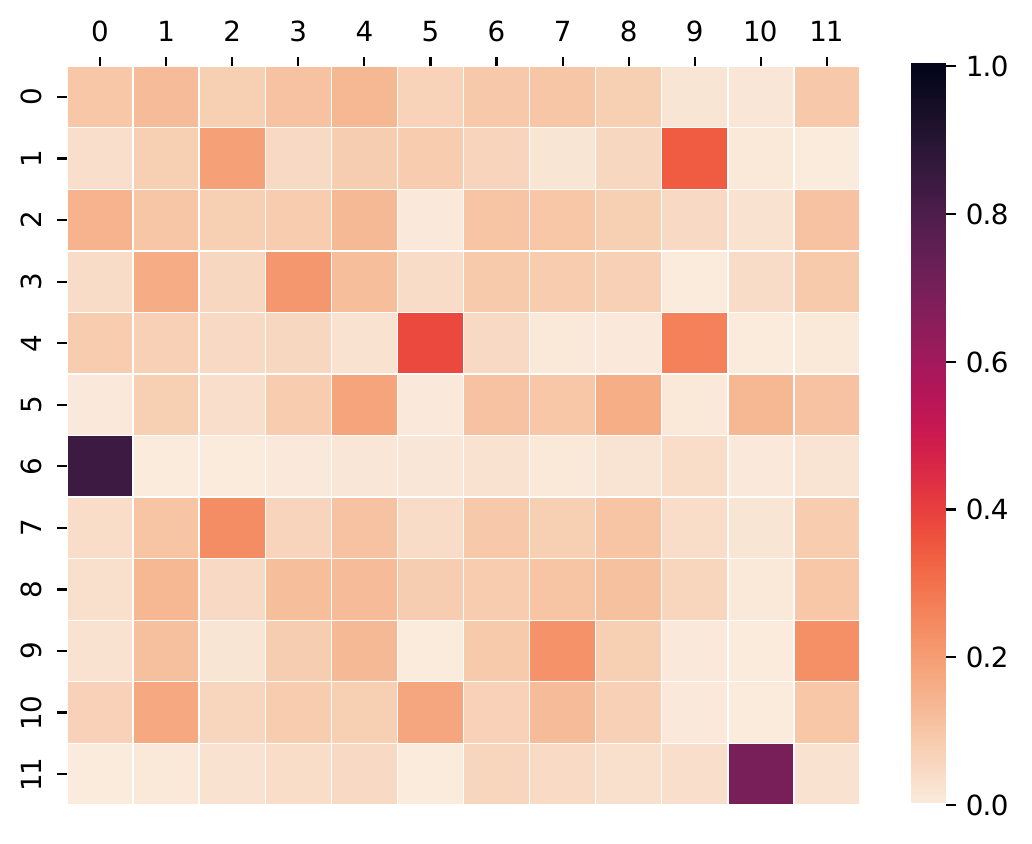}
        \caption{SG Co-Opt, Section~\ref{sect:coopt}}
        \label{fig:SG-coopt}
     \end{subfigure}
        \caption{Heatmaps of optimized single-robot patrol strategies for San Francisco graph from RoSSO.}
        \label{fig:SF_heatmaps}
\end{figure*}

Now we compare the MC metrics presented in Section~\ref{sect:problem_form}. Fig.~\ref{fig:SF_heatmaps} shows the best results obtained from running RoSSO for the MHT, SG, and RTE formulations with 100 random initializations. The corresponding objective function and penalty values are given in Table~\ref{table:forms}.
% We used a Google Cloud virtual machine with an NVIDIA L4 GPU and 32 Intel Xeon Scalable vCPUs. 
$\alpha = 1$ was used in the penalty term of the objective function. The stopping criterion for the optimization was a mean relative change in objective function value less than 0.01 over 10 successive iterations. 
% The iteration limit was set to $1\mathrm{e}{4}$. 
The \texttt{RMSprop} gradient-based optimization algorithm was found empirically to be effective on these problems~\cite{TT-GH:12}. We also discovered that convergence occurs more quickly for the SG formulation if the mean of the lowest 4 capture probabilities is maximized rather than just the lowest. The computational efficiency data is given in Table~\ref{table:comp_data} and shows that the RTE formulation is much more computationally expensive than the other formulations. 
% Note that the presence of heterogeneous travel times necessitates the use of~\eqref{eq:F_k_het} for computing the first hitting time matrices $F_k$ for the SG and RTE formulations. The MHT formulation appears slower on this hardware compared to the data in Table~\ref{table:HW} because a key subroutine has yet to be implemented in JAX for GPU hardware.

We see in Fig.~\ref{fig:MHT} that the MHT strategy is asymmetric and sparse. This strategy can be thought of as a cycle over the graph with some probability of staying at intersections 0 and 3 for multiple consecutive time periods. This allows the strategy to meet the stationary distribution constraint. A robot patroller using this strategy will be quite predictable. 
% As discussed in~\cite{XD-FB:20a}, imposing reversibility on the Markov chain for convexity of the optimization leads to a symmetric strategy with a weighted mean hitting time $\M^w = 44.77$, a severe decrease in performance. Therefore, using RoSSO with multiple random initializations is a practical way to find fast-moving patrol strategies. 
The RTE strategy shown in Fig.~\ref{fig:RTE} is asymmetric and dense. This strategy resembles a uniform random walk that meets the stationary distribution constraint. Clearly, a robot following this patrol strategy will be hard to predict. Note that the truncation accuracy was set to $\eta = 0.1$ as in~\cite{XD-MG-FB:17o}. This corresponds to $K_\eta = 2295$ in Eq.~\eqref{eq:F_k_het} and explains the slow computation shown in Table~\ref{table:comp_data}. 
The SG strategy shown in Fig.~\ref{fig:SG} is asymmetric and a middle ground in terms of sparsity. This strategy has features similar to both the MHT and the RTE strategies. The vector of attack durations $\tauvec = 9 \cdot \1_n$ was chosen as the worst-case uniform attack duration scenario with a nonzero capture probability.
\begin{table*}
\centering
\caption{Averaged computational efficiency data for RoSSO.}
\label{table:comp_data}
\begin{tabular}{| p{2.5cm} | p{0.75cm} p{0.75cm} p{0.75cm} p{1.5cm} | p{1.25cm} p{1.25cm} | p{1.75cm} p{1.75cm} |} 
 \hline
  & MHT & RTE & SG & SG Co-Opt & Multi-SG ($R=2$) & Multi-SG ($R=3$) & Part.~Multi-SG ($R=2$) & Part.~Multi-SG ($R=3$) \\
 \hline
 Avg. Wall Time [s] & 1.14 & 11.5 & 3.07 & 17.2 & 10.9 & 174 & 1.81 & 1.72 \\ 
 Avg. No. of Iterations & 460 & 12 & 2103 & 2697 & 2577 & 1431 & 444 & 209 \\
 Avg. Speed [iter/s] & 405 & 1.04 & 685 & 156 & 237 & 8.24 & 245 & 122 \\
 \hline
\end{tabular}
\end{table*}

% \subsection{Stackelberg Game Formulation: Co-Optimizing Defense Placement \& Patrol Strategy}
Now we consider the simultaneous optimization of defense placement and patrol strategy in the SG formulation as described in Section~\ref{sect:coopt}. The defense budget $B = 9 \times 12 = 108$ to compare with the results in Fig.~\ref{fig:SG}. Fig.~\ref{fig:SG-coopt} displays the best resulting patrol strategy from 100 random initializations. The corresponding vector of attack durations $\tauvec = \begin{bmatrix}
    8 & 6 & 11 & 10 & 6 & 10 & 9 & 10 & 11 & 9 & 10 & 8
\end{bmatrix}^\top$. Note that this nonuniform choice of $\tauvec$ leads to a relative increase of 5\% in SG capture probability as shown in Table~\ref{table:forms}. The drawback is increased computational expense as shown in Table~\ref{table:comp_data}.

Table~\ref{table:forms} gives the values of all three metrics for each of the optimized strategies shown in Fig.~\ref{fig:SF_heatmaps}. Recall that the aim is to minimize $J_{\textup{MHT}}$, Penalty and maximize $J_{\textup{RTE}}, J_{\textup{SG}}$. We see that the MHT strategy performs poorly when evaluated by the RTE and SG metrics. This is due to the sparsity of the strategy and the long intervals between departure and arrival for certain pairs of nodes, respectively. The RTE strategy suffers on the MHT metric due to slow traversal of the patrol area. The SG strategy and the co-optimized SG strategy perform reasonably well on all metrics. The $J_{\textup{MHT}}$ values 48.8 and 47.9 are acceptable considering that the optimal value is 44.8 when including the reversibility constraint. 
\begin{table}[H]
\centering
\caption{Evaluation of the optimized strategies from Fig.~\ref{fig:SF_heatmaps} by each metric.}
\label{table:forms}
\begin{tabular}{| c | c c c c |} 
 \hline
  & Fig.~\ref{fig:MHT} & Fig.~\ref{fig:RTE} & Fig.~\ref{fig:SG} & Fig.~\ref{fig:SG-coopt} \\
 \hline
 $J_{\textup{MHT}}$ & 21.2 & 58.1 & 48.8 & 47.9 \\ 
 $J_{\textup{RTE}}$ & 2.10 & 5.13 & 4.71 & 4.64 \\
 $J_{\textup{SG}}$ & $1.17\mathrm{e}{-8}$ & $1.97\mathrm{e}{-2}$ & $9.75\mathrm{e}{-2}$ & $10.2\mathrm{e}{-2}$ \\
 Penalty & $9.94\mathrm{e}{-3}$ & $2.97\mathrm{e}{-2}$ & $5.13\mathrm{e}{-3}$ & $4.70\mathrm{e}{-3}$ \\
 \hline
\end{tabular}
\end{table}

\subsection{Multi-Robot Patrolling} \label{sect:res_multi}
\begin{figure*}
    \centering
     \begin{subfigure}[b]{0.19\textwidth}
         \centering
         \includegraphics[width=\textwidth]{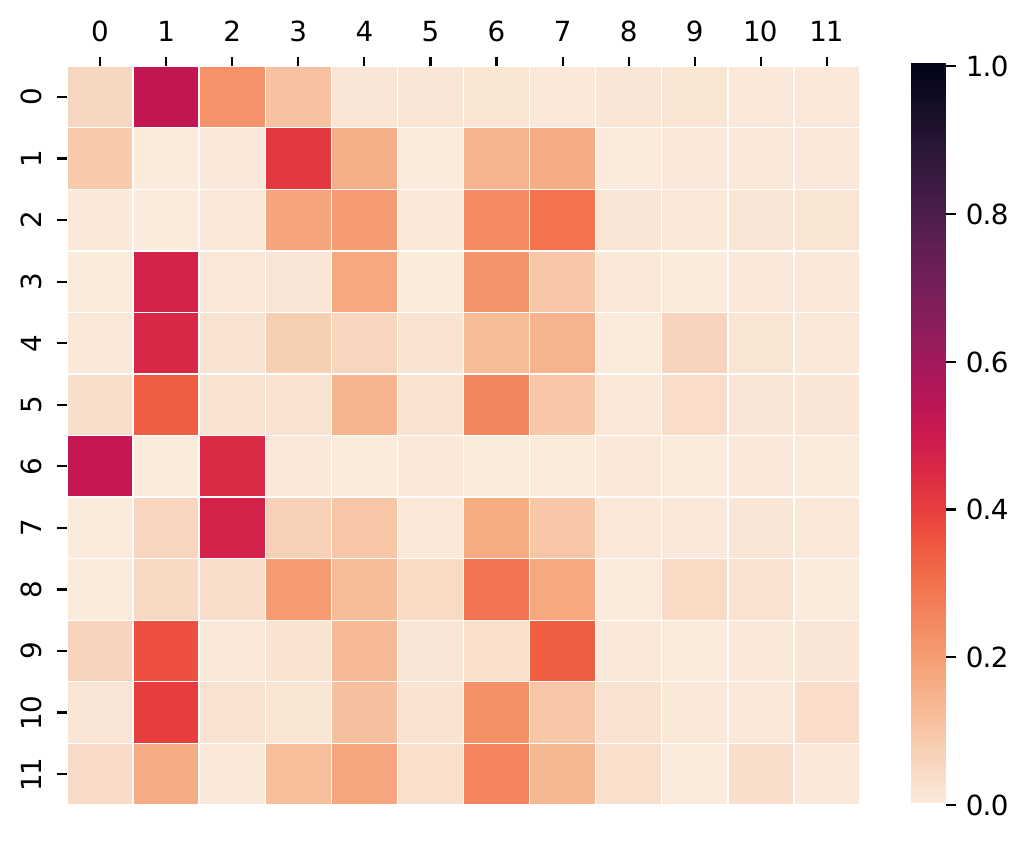}
         \caption{$R = 2$, \\ $J_{\textup{SGM}} = 0.167$, \\ $\text{penalty} = 0.664$}
         \label{fig:multi_N2a}
     \end{subfigure}
     \hfill
     \begin{subfigure}[b]{0.19\textwidth}
         \centering
         \includegraphics[width=\textwidth]{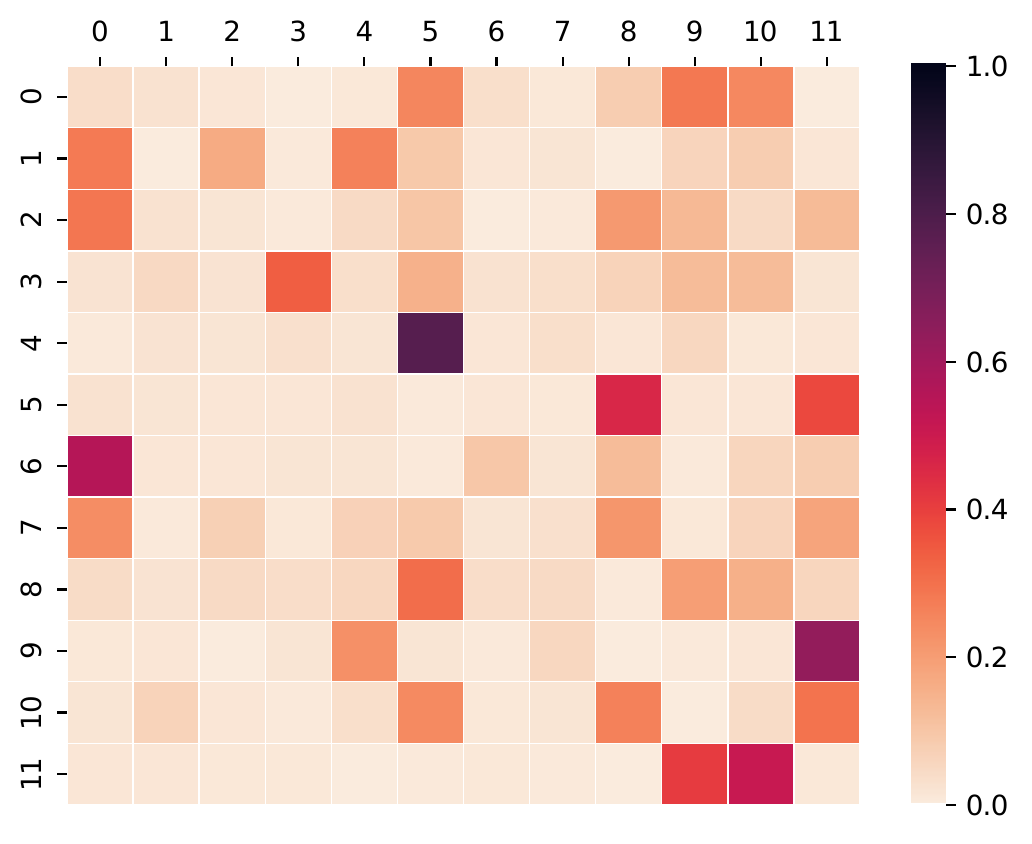}
         \caption{$R = 2$, \\ $J_{\textup{SGM}} = 0.167$, \\ $\text{penalty} = 0.664$}
         \label{fig:multi_N2b}
     \end{subfigure}
     \hfill
     \begin{subfigure}[b]{0.19\textwidth}
         \centering
         \includegraphics[width=\textwidth]{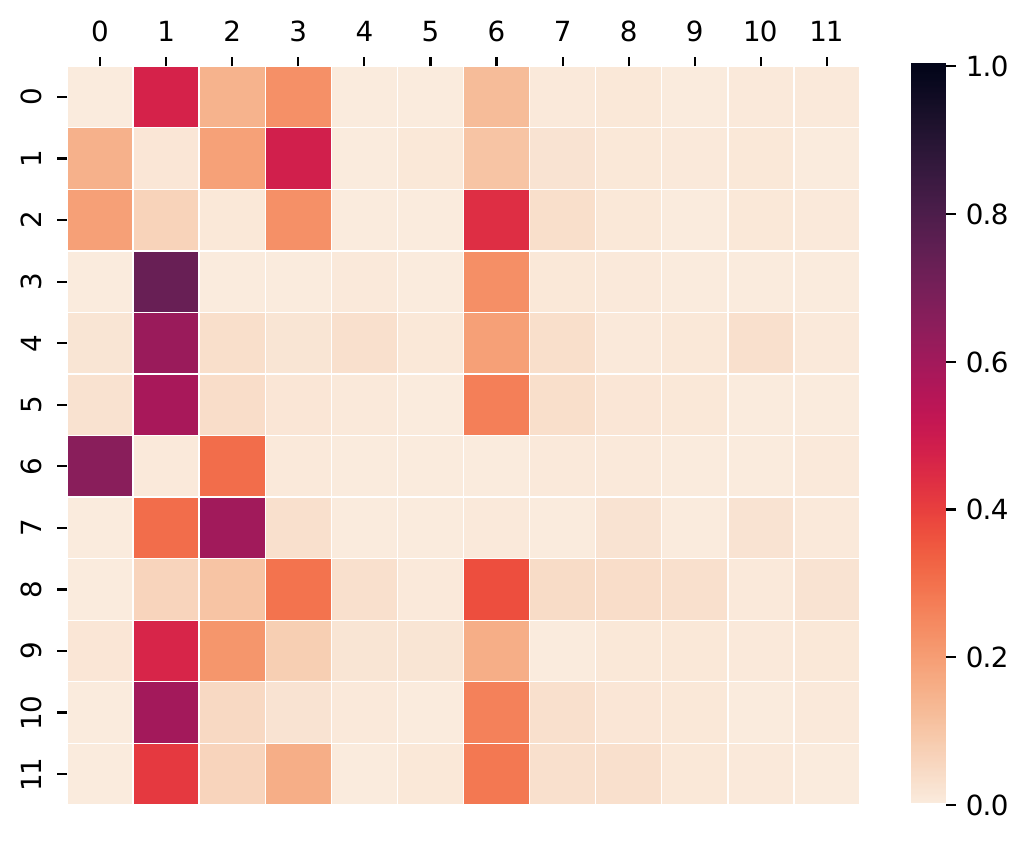}
         \caption{$R = 3$, \\ $J_{\textup{SGM}} = 0.278$, \\ $\text{penalty} = 0.207$}
         \label{fig:multi_N3a}
     \end{subfigure}
     \hfill
     \begin{subfigure}[b]{0.19\textwidth}
         \centering
         \includegraphics[width=\textwidth]{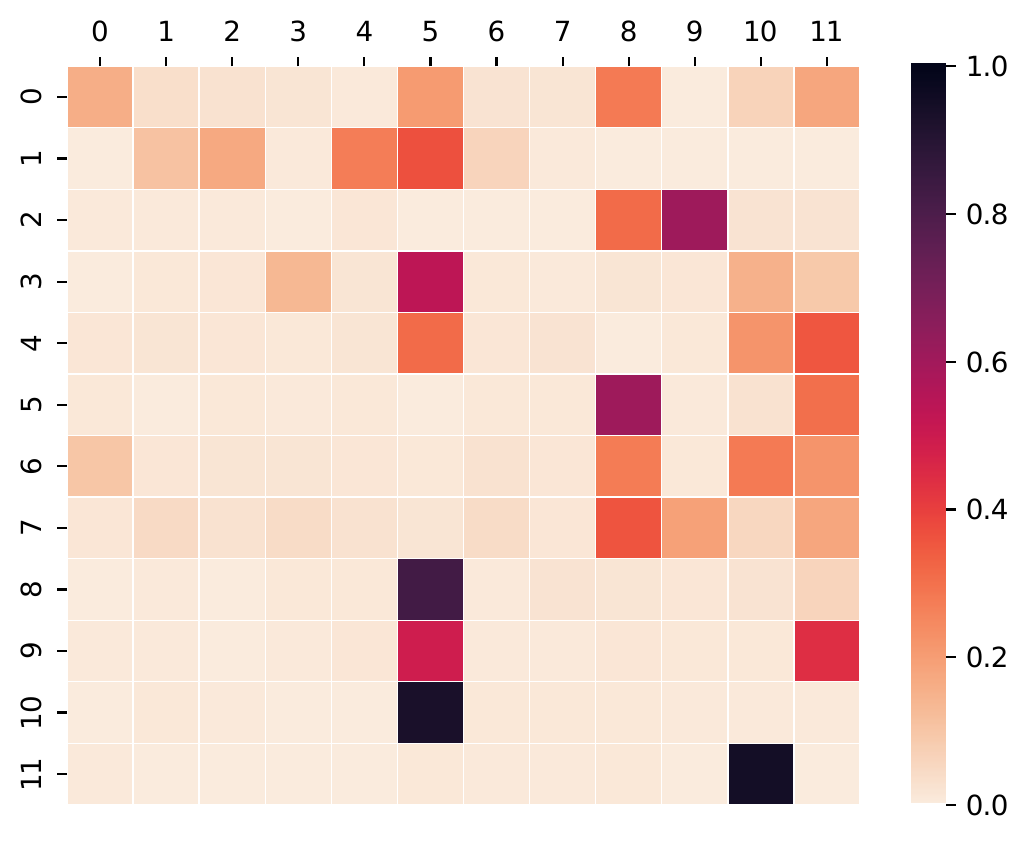}
         \caption{$R = 3$, \\ $J_{\textup{SGM}} = 0.278$, \\ $\text{penalty} = 0.207$}
         \label{fig:multi_N3b}
     \end{subfigure}
     \hfill
     \begin{subfigure}[b]{0.19\textwidth}
         \centering
         \includegraphics[width=\textwidth]{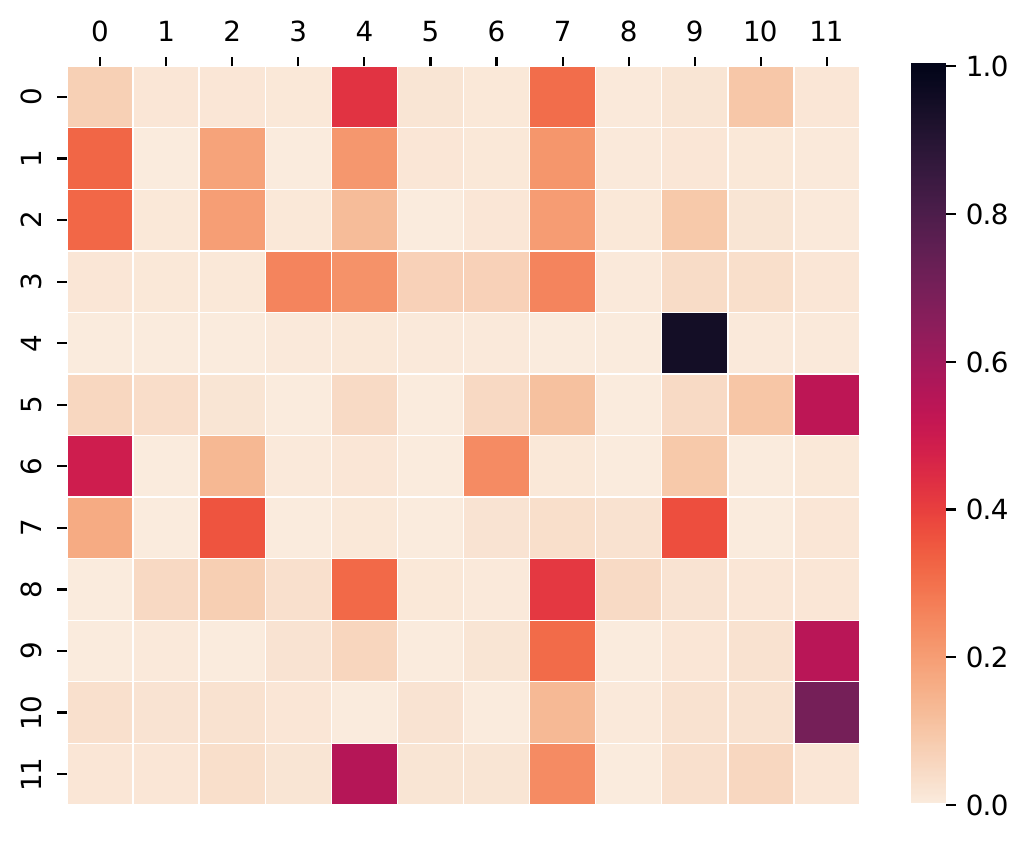}
         \caption{$R = 3$, \\ $J_{\textup{SGM}} = 0.278$, \\ $\text{penalty} = 0.207$}
         \label{fig:multi_N3c}
     \end{subfigure}
     \caption{Heatmaps of SG-optimized robot team patrol strategies for San Francisco graph from RoSSO, Section~\ref{sect:multi}.}
     \label{fig:SF_multi_heatmaps}
\end{figure*}

Here we compare our proposed multi-robot SG formulation from Section~\ref{sect:multi} with a partitioning approach. We focus on the SG metric due to the aforementioned balanced nature of the SG strategies. We obtained better results for this formulation by maximizing just the lowest capture probability. The other parameters are unchanged from the single-robot results. Fig.~\ref{fig:SF_multi_heatmaps} shows the best optimized strategies from 10 random initializations for teams of robot patrollers. Figs.~\ref{fig:multi_N2a},~\ref{fig:multi_N2b} contain the $R = 2$ results, and Figs.~\ref{fig:multi_N3a},~\ref{fig:multi_N3b},~\ref{fig:multi_N3c} contain the $R = 3$ results. We can see that the optimized strategies naturally divide up the graph and focus each robot's attention on a subset of the intersections. We emphasize that this is purely emergent behavior; no restrictions were imposed beyond MC validity and the stationary distribution penalty given in Eq.~\eqref{eq:multi_penalty}. For example in the $R = 2$ results, the first robot is biased towards intersections 0-4, 6, and 7 while the second robot is biased towards intersections 5 and 8-11. Examining the map in Fig.~\ref{fig:SF}, we see that this is a reasonable division of labor.
The computational efficiency data in Table~\ref{table:comp_data} shows the expected drastic slowdown as the number of robots $R$ increases.

\begin{figure*}
    \centering
     \begin{subfigure}[b]{0.19\textwidth}
         \centering
         \includegraphics[width=\textwidth]{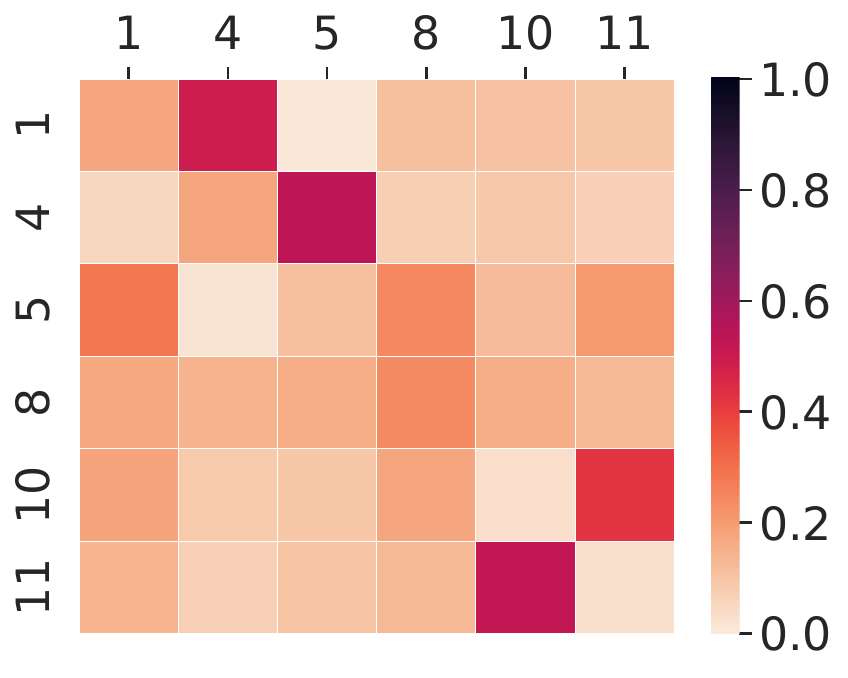}
         \caption{$R = 2$, \\ $J_{\textup{SG}} = 0.307$, \\ $\text{penalty} = 1.26\mathrm{e}{-2}$}
         \label{fig:part_N2a}
     \end{subfigure}
     \hfill
     \begin{subfigure}[b]{0.19\textwidth}
         \centering
         \includegraphics[width=\textwidth]{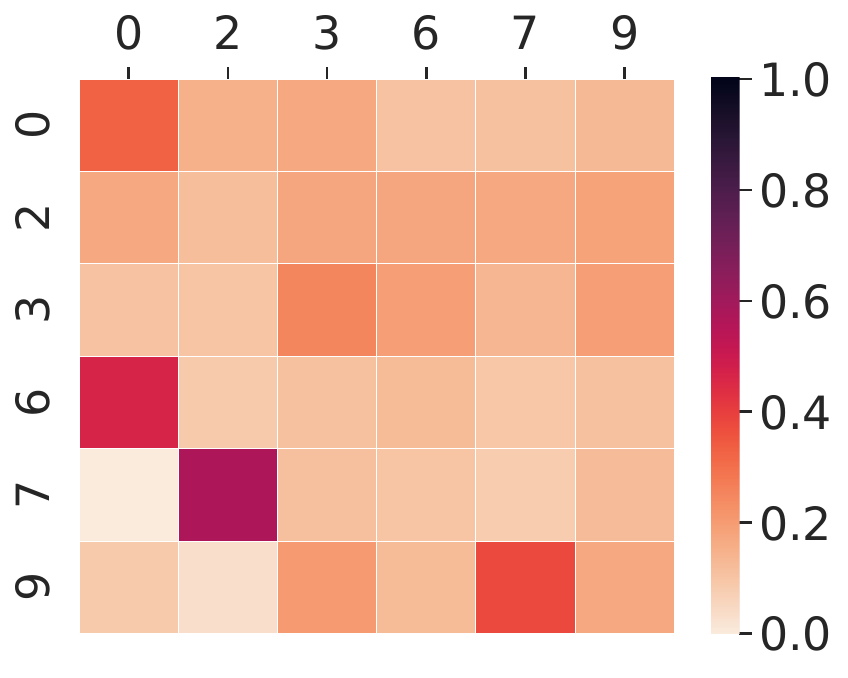}
         \caption{$R = 2$, \\ $J_{\textup{SG}} = 0.260$, \\ $\text{penalty} = 7.43\mathrm{e}{-3}$}
         \label{fig:part_N2b}
     \end{subfigure}
     \hfill
     \begin{subfigure}[b]{0.19\textwidth}
         \centering
         \includegraphics[width=\textwidth]{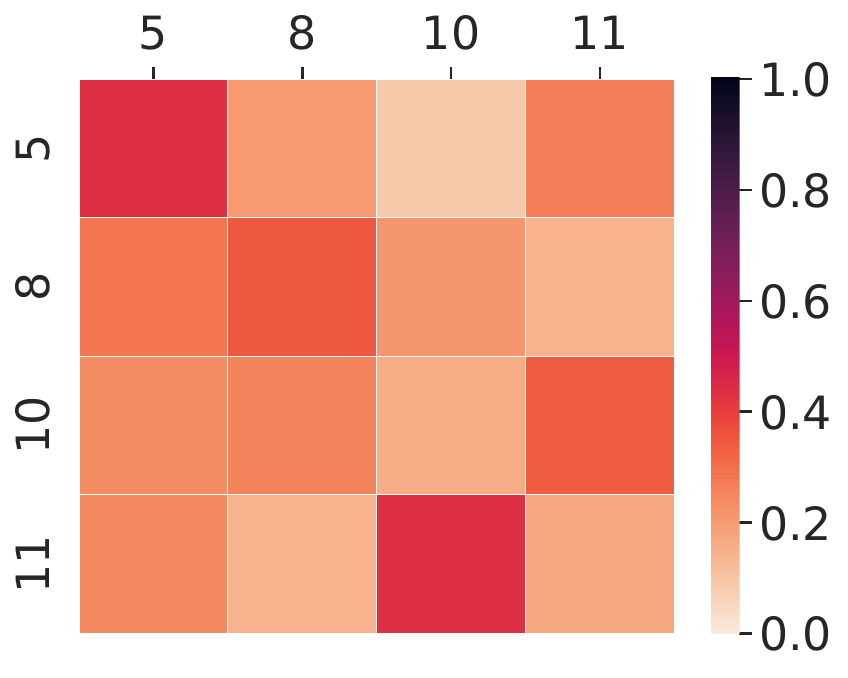}
         \caption{$R = 3$, \\ $J_{\textup{SG}} = 0.488$, \\ $\text{penalty} = 1.06\mathrm{e}{-2}$}
         \label{fig:part_N3a}
     \end{subfigure}
     \hfill
     \begin{subfigure}[b]{0.19\textwidth}
         \centering
         \includegraphics[width=\textwidth]{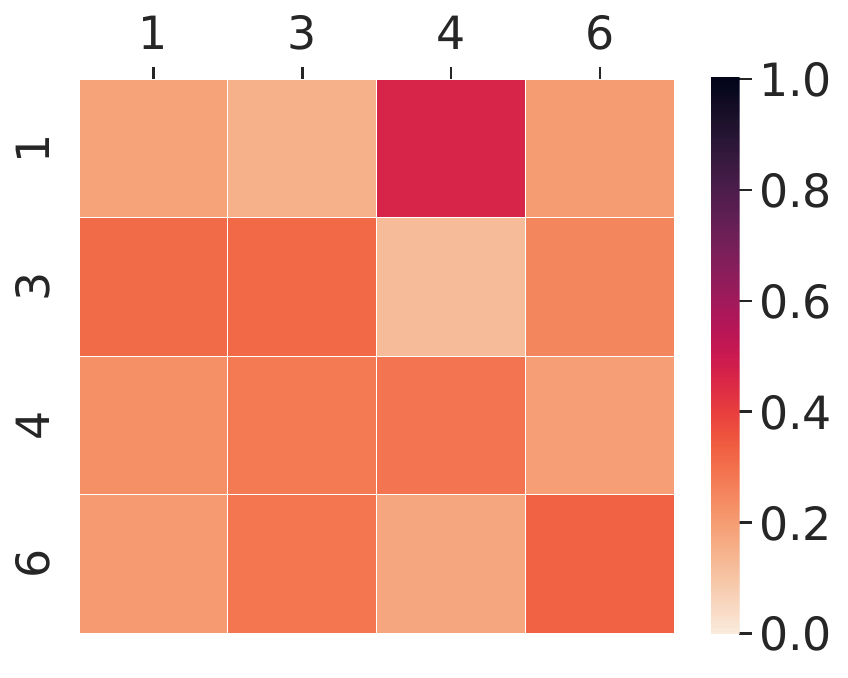}
         \caption{$R = 3$, \\ $J_{\textup{SG}} = 0.416$, \\ $\text{penalty} = 2.06\mathrm{e}{-3}$}
         \label{fig:part_N3b}
     \end{subfigure}
     \hfill
     \begin{subfigure}[b]{0.19\textwidth}
         \centering
         \includegraphics[width=\textwidth]{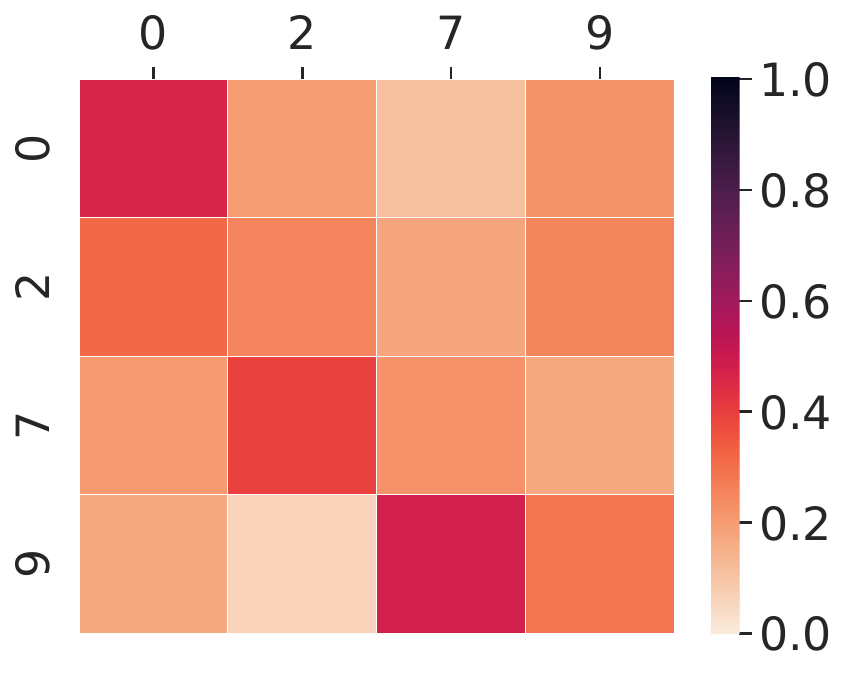}
         \caption{$R = 3$, \\ $J_{\textup{SG}} = 0.514$, \\ $\text{penalty} = 1.59\mathrm{e}{-2}$}
         \label{fig:part_N3c}
     \end{subfigure}
        \caption{Heatmaps of SG-optimized robot team patrol strategies for partitioned San Francisco graph from RoSSO.}
        \label{fig:SF_part_heatmaps}
\end{figure*}

We compare with a prescribed partition approach. The partitions are chosen a priori as shown in Fig.~\ref{fig:SF}. Fig.~\ref{fig:SF_part_heatmaps} shows the best optimized strategies from 10 random initializations for single-robot patrollers confined to partitions of the graph. Figs.~\ref{fig:part_N2a},~\ref{fig:part_N2b} contain the $R = 2$ results, and Figs.~\ref{fig:part_N3a},~\ref{fig:part_N3b},~\ref{fig:part_N3c} contain the $R = 3$ results. Notice that the capture probabilities are higher with this approach. However, this depends on appropriate partitioning which is a nontrivial problem for large graphs. Additionally, the set of possible strategies when using a partitioning approach is a subset of the strategies that can be described by the un-partitioned MCs. With a sufficient number of random initializations, we expect the un-partitioned approach to achieve equivalent or better performance.
The computational efficiency data in Table~\ref{table:comp_data} shows the appeal of the partitioning approach. Again, we emphasize that additional computation time would be needed for choosing appropriate partitions in a large graph. We also remark that the first iteration of optimization using RoSSO can take noticeably longer than following iterations due to accelerated linear algebra compilation. This can slightly skew the average speed figures presented in Table~\ref{table:comp_data} when the number of iterations is small.

\section{Conclusions}
RoSSO is a valuable tool for enabling MC optimization over a realistic graph with travel times and varying node priority. Using RoSSO, we conducted a case study on the San Francisco crime graph that identified the SG metric as a good choice for generating patrol strategies that balance movement speed and unpredictability. Focusing on the SG formulation, we proposed a novel greedy algorithm for the design problem of co-optimizing defense budget allocation and patrol strategy. On the multi-robot front, we proposed a novel SG formulation and leveraged the computational efficiency of RoSSO to optimize it. We compared against a graph partitioning approach and discovered that appropriately chosen partitions can narrow the state space and lead to effective strategies with fewer random initializations and less computational expense per initialization. In future, we aim to achieve further speedup via parallelized computation on GPU/TPU hardware. Additionally, the JAX community has developed several reinforcement learning libraries that we intend to explore for robotic surveillance applications.
% Additionally, reinforcement learning offers an alternative approach to the one discussed here, and JAX has several RL libraries that could be brought to bear.


%%%%%%%%%%%%%%%%%%%%%%%%%%%%%%%%%%%%%%%%%%%%%%%%%%%%%%%%%%%%%%%%%%%%%%%%%%%%%%%%

\section*{Appendix}
Here we define the stationary distribution and first hitting times of an MC. The MC $P$ is irreducible if every node is reachable from every other node. An irreducible MC $P$ has a unique nonnegative stationary distribution $\pivec \in \real^{n}$ that satisfies $\pivec^\top P = \pivec^\top$, $\pivec^\top \1_n = 1$. The stationary distribution represents the fraction of the robot's time spent at each node in the graph. 
Let $X_k\in \{1,\ldots,n\}$ be the state of an MC at time period $k$. The first hitting time, $T_{ij}$, is a random variable representing the duration between the robot's departure from waypoint $i$ and its \textit{first} arrival at waypoint $j$, i.e., $T_{ij} = \min \{k \mid X_0 = i, X_k = j, k \ge 1\}$. 
% A quantity of interest is the first hitting time denoted $T_{ij}$ and defined as a random variable that represents the number of time periods between the robot leaving waypoint $i$ and its \textit{first} arrival at waypoint $j$, i.e., $T_{ij} = \min \{k \mid X_0 = i, X_k = j, k \ge 1\}$. 
The first hitting time probability matrices $F_k$ can be defined wherein $F_k(i,j) = \Prob(T_{ij} = k)$. 
% Therefore, there is a probability associated with each value of $k \ge 1$. These first hitting time probabilities can be collected into a sequence of matrices $F_k$ where $F_k(i,j) = \Prob(T_{ij} = k)$. 
Recursions for computing $F_k$ are presented in~\cite{XD-MG-FB:17o}. For graphs with homogeneous travel times, we have the following where $F_1 = P$:
\begin{equation} \label{eq:F_k_homo}
    \vecvec(F_k) = (I_n \otimes P) (I_{n^2} - \diag(\vecvec(I_n))) \vecvec(F_{k-1}),
\end{equation}
For graphs with heterogeneous travel times, we have:
\begin{equation} \label{eq:F_k_het}
\begin{aligned}
    & \vecvec(F_k) = \vecvec \bigl( P \circ \bm{1}_{\{k \1_n\1_n^\top = W\}} \bigr) \\
    & + \sum\nolimits_{i=1}^n \sum\nolimits_{j=1}^n P(i,j) \bigl( E_j \otimes \e_i \e_j^\top \bigr) \vecvec \bigl( F_{k-w_{ij}} \bigr),
\end{aligned}
\end{equation}
where $\bm{1}_{\{ \cdot \}}$ is the indicator function, $E_j = \diag (\1_n - \e_j)$, and $F_k = \0_{n \times n}$ for all $k \leq 0$.

% \section*{Acknowledgment}
% The authors thank Sean Jaffe and Max Emerick for valuable discussions in the early stages of this work.

%%%%%%%%%%%%%%%%%%%%%%%%%%%%%%%%%%%%%%%%%%%%%%%%%%%%%%%%%%%%%%%%%%%%%%%%%%%%%%%%

% \FloatBarrier
\clearpage

\bibliographystyle{plain}
\bibliography{bib/alias, bib/FB, bib/Main, bib/YJ}

\end{document}